\documentclass[journal]{IEEEtran}

\usepackage{colortbl}  
\usepackage{xcolor}    
\usepackage[utf8]{inputenc} 
\DeclareUnicodeCharacter{03BC}{\ensuremath{\mu}} 
\usepackage[table]{xcolor}
\usepackage{booktabs}
\usepackage{tabularx}
\usepackage{array}
\newcolumntype{P}[1]{>{\centering\arraybackslash}p{#1}}
\usepackage{comment}
\usepackage[version=4]{mhchem}
\usepackage{xcolor}
\usepackage{soul}
\usepackage{soul} 
\usepackage{soul}
\usepackage{xcolor}
\usepackage[utf8]{inputenc}
\usepackage{upgreek}

\usepackage{cite}
\usepackage{siunitx}
\usepackage{graphicx}
\usepackage{caption}
\usepackage{hyperref}        

\begin{document}
\title{Micro-Dexterity in Biological Micromanipulation: Embodiment, Perception, and Control }

\author{
Kangyi Lu, Lan Wei, Zongcai Tan, Dandan Zhang
\thanks{All authors are with the Department of Bioengineering, Imperial-X AI Initiative, Imperial College London, London, United Kingdom.  Corresponding: d.zhang17@imperial.ac.uk.}
}

\maketitle

%

\begin{abstract}

Microscale manipulation has advanced substantially in controlled locomotion and targeted transport, yet many biomedical applications require precise and adaptive interaction with biological micro-objects. At these scales, manipulation is realized through three main classes of platforms: embodied microrobots that physically interact as mobile agents, field-mediated systems that generate contactless trapping or manipulation forces, and externally actuated end-effectors that interact through remotely driven physical tools.  Unlike macroscale manipulators, these systems function in fluidic, confined, and surface-dominated environments characterized by negligible inertia, dominant interfacial forces, and soft, heterogeneous, and fragile targets. Consequently, classical assumptions of dexterous manipulation, including rigid-body contact, stable grasping, and rich proprioceptive feedback, become difficult to maintain.
This review introduces micro-dexterity as a framework for analyzing biological micromanipulation through the coupled roles of embodiment, perception, and control. We examine how classical manipulation primitives, including pushing, reorientation, grasping, and cooperative manipulation, are reformulated at the microscale; compare the architectures that enable them, from contact-based micromanipulators to contactless field-mediated systems and cooperative multi-agent platforms; and review the perception and control strategies required for task execution. We identify the current dexterity gap between laboratory demonstrations and clinically relevant biological manipulation, and outline key challenges for future translation.

\end{abstract}

\section{Introduction}

Microrobots operating at cellular and tissue scales constitute an emerging platform with significant potential to enable new paradigms in biomedical intervention \cite{zhang2023advanced}. By integrating actuation, mobility, and control within the micrometer regime, these systems could advance targeted drug delivery, single-cell analysis, and minimally invasive surgical procedures \cite{dabbagh20223d, hou2023review, lin2024magnetic}. Early demonstrations have established that microrobots can navigate fluidic environments and perform tasks such as cargo transport and deployment \cite{landers2025clinically}. However, as the field transitions from proof-of-concept mobility to complex biological applications, the demand for operational capability is shifting toward micro-dexterous manipulation, encompassing primitives such as pushing, reorientation, grasping, and cooperative manipulation of biological objects.

In the realm of large-scale robotics, dexterous manipulation is defined as the ability of an end-effector to alter the position and orientation of an object through precise, coordinated movements and adaptive force modulation \cite{okamura2000overview}. It extends beyond simple pick-and-place operations to include complex activities like tool use, in-hand object reorientation, and fine assembly. Achieving such dexterity requires sophisticated feedback control to handle high-dimensional motion planning and multi-contact dynamics \cite{yu2022dexterous, an2025dexterous}.


Rather than a straightforward exercise in miniaturization, translating macroscale dexterous capabilities to the microscale is fundamentally constrained by multiple intersecting scaling laws. 
First, the physical regime changes fundamentally in confined fluidic environments, with negligible inertia and dynamics dominated by viscous drag and surface adhesion \cite{Zhang2016Fundamentals}.  Second, target properties present unique challenges, as biological objects are soft, heterogeneous, and fragile, imposing strict limits on contact forces to preserve viability. Third, embodiment limits preclude the use of complex multi-degree-of-freedom (DoF) linkages and onboard sensors, necessitating indirect, globally coupled actuation via external fields \cite{alu2025roadmap}. Finally, perception constraints arise from the absence of proprioception, forcing a reliance on external imaging that hinders state estimation \cite{gao2025soft}. 
Consequently, micro-dexterity is commonly realized through alternative paradigms, including functional material intelligence in the form of embodied compliance and stimuli responsiveness \cite{tanembodied}, collective multirobot coordination, and programmable field-mediated interactions \cite{chen2025physical}.

\begin{figure*}
    \centering
    \captionsetup{font=footnotesize,labelsep=period}
    \includegraphics[width=0.95\linewidth]{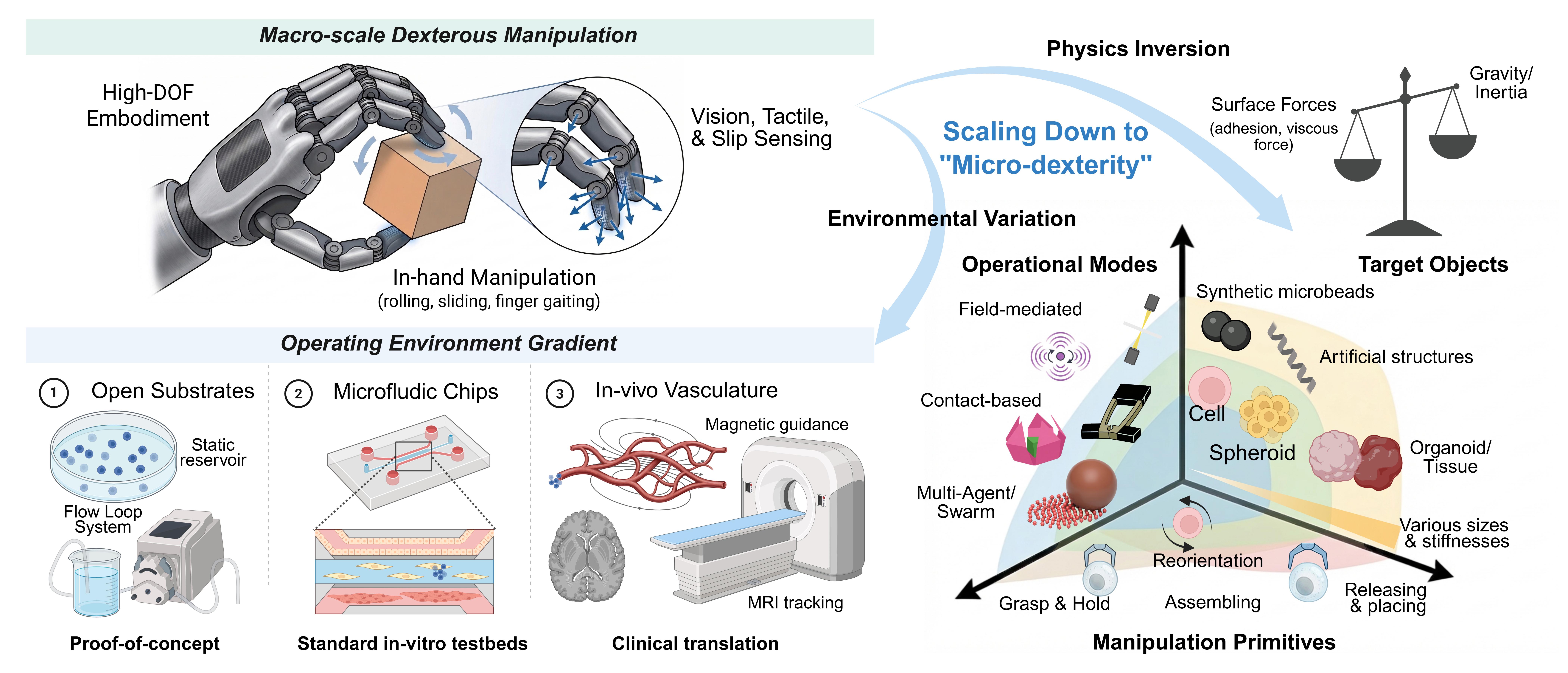}
    \caption{From macroscale manipulation to micro-dexterity capability space. Upper left: macroscale dexterity with high-DOF embodiment and multimodal sensing. During downscaling, two fundamental shifts emerge: physics inversion from inertia-dominated to adhesion-dominated interactions, and an environment gradient from open substrates to in vivo vasculature (lower left). Right: the multidimensional capability space of manipulation primitives, target objects, and operational modes.}
    \label{fig:overview}
    \vspace{-0.5cm}
\end{figure*} 

While the microrobotics literature continues to expand, recent reviews have largely organized the field around a “materials–fabrication–actuation–control–translation” pipeline. This perspective has clarified how microrobotic systems are realized through specific materials and fabrication strategies \cite{dabbagh20223d, chen2025physical, chen2024materials}, how they achieve locomotion under different propulsion modalities and environmental constraints \cite{hou2023review, hussein2023actuation}, how they are governed through feedback-based tracking and autonomous control \cite{jiang2022control}, and which application domains they address \cite{wang2024soft}. However, such a component-wise organization also encourages a fragmented view, in which progress is assessed primarily within individual technical axes. What remains less well understood is how these elements jointly enable task-adaptive interaction with biological targets.

To address this gap, we introduce \textbf{`micro-dexterity'} as the capability of microrobotic systems to construct, regulate, and adapt mechanical interactions with biological targets so as to reliably alter their state under microscale conditions. Unlike macroscale dexterity, which often depends on fully actuated multi-DOF end-effectors and rich proprioceptive feedback, micro-dexterity is shaped by low-Reynolds-number dynamics, adhesion-dominated interactions, stringent force limits imposed by soft living matter, and globally coupled actuation. It should therefore be assessed not only by the execution of isolated manipulation primitives, but by how reliably such primitives can be composed into adaptive behaviors in biological environments.

This review adopts a capability-oriented perspective on biological micromanipulation. Rather than providing an exhaustive survey, it offers an expert-curated synthesis of representative and influential studies across four coupled dimensions: embodiment and actuation, sensing and perception, control and learning, and task-level capabilities. We prioritize systems involving biological targets or biologically relevant constraints, while incorporating non-biological examples only when they reveal transferable mechanisms of dexterous interaction. By organizing the literature around the emerging concept of micro-dexterity, this review complements existing component-centered surveys and provides an integrated framework for understanding how advances in embodiment, perception, and control collectively enable biological micromanipulation.

This review first reformulates classical notions of dexterity under microscale physical and biological constraints (Section II). We then examine the embodiment and actuation principles that realize microscale manipulation, with emphasis on how materials, morphology, and field-mediated interactions compensate for the absence of conventional articulated linkages (Section III). Next, we consider the shift from direct physical sensing to exteroceptive perception and physics-informed state estimation as the basis for feedback at small scales (Section IV), followed by the control and learning strategies required to transform low-dimensional external actuation into adaptive manipulation behaviors (Section V). Finally, we synthesize these elements through task-level analysis, highlighting the current “dexterity gap” and identifying key milestones for clinical translation (Sections VI and VII).

\section{From Macro to Micro Dexterous Manipulation}

\subsection{Dexterous Manipulation at the Macro Scale}
In classical robotics, dexterous manipulation at the macroscale refers to a robot hand’s or end-effector’s ability to regulate multi-contact interactions so as to reposition and reorient objects, reconfigure contact states, and execute in-hand or tool-mediated actions \cite{bicchi2002hands}. Such capability is typically enabled by kinematic redundancy and controllable internal forces, together with feedback mechanisms for contact-state estimation, slip detection, and force modulation during contact-rich motion \cite{sampath2023review,fan2025magicgripper}. 

Recent progress has been driven by the co-development of compliant multi-DOF robotic hands \cite{piazza2019century}, tactile and vision-based perception \cite{fan2025crystaltac}, and learning-enabled control frameworks \cite{ye2026visual}. Tactile sensing provides local measurements of contact geometry and shear interaction, while proprioceptive and exteroceptive feedback are increasingly integrated to improve robustness under uncertainty \cite{xia2022review, an2025dexterous}. At the same time, reinforcement and imitation learning have enabled increasingly capable contact-rich behaviors, including in-hand manipulation, precision assembly, and coordinated multi-arm action \cite{chen2023visual, luo2025precise}, with physics-based simulation serving as a key enabler for large-scale training, benchmarking, and sim-to-real transfer \cite{todorov2012mujoco, makoviychuk2021isaac}.

Overall, macroscale dexterous manipulation is characterized by high-DOF embodiments, multimodal sensing, and closed-loop control for robust contact management. However, extending this paradigm to the microscale is not a simple exercise in miniaturization, because fundamental changes in targets, environments, and interaction physics require dexterity to be reformulated under microscale conditions.

\subsection{Physical Constraints and Opportunities at the Microscale}

As manipulation scales down to the microscale, the governing physics changes fundamentally, creating constraints and opportunities that differ from those at the macroscale. Micromanipulation typically occurs in a low-Reynolds-number regime, where viscous drag dominates inertia and net motion requires continuous non-reciprocal actuation \cite{ppurcell2014life}. At the same time, scaling laws amplify surface forces relative to gravity, leading to the ``easy-to-grasp, difficult-to-release'' challenge and making reliable detachment a central issue in micro-object handling \cite{chang2021capillary}. These difficulties become even more substantial for biological targets, whose low stiffness and viscoelasticity demand compliant and carefully regulated interactions to avoid mechanical damage.

These physical constraints manifest differently across the diverse operating environments in which microrobots are deployed. Open substrates offer well-controlled conditions for proof-of-concept studies but only partially reflect realistic deployment scenarios \cite{bozuyuk2024roadmap}. Microfluidic chips introduce bounded geometries and controlled flow profiles, serving as standardized in vitro testbeds for evaluating transport and manipulation under bounded conditions \cite{dong2024ai}. In vivo vasculature presents the most demanding setting, where pulsatile blood flow, non-Newtonian rheology, tortuous geometries, and immune clearance drastically alter microrobot locomotion and require real-time medical imaging for closed-loop guidance \cite{del2023ultrasound, li2024human}. This environmental gradient from static substrates to microfluidic channels and living vasculature, places progressively greater demands on actuation, feedback, and biocompatibility, necessitating environment-specific manipulation strategies. 

At the same time, the microscale regime also creates opportunities to exploit dominant environmental physics. Surface tension, electrostatic interactions, and viscous drag can be harnessed for gripping, trapping, and gentle steering \cite{seon2017enhance}. As a result, microscale manipulation increasingly departs from traditional rigid-linkage architectures in favor of alternative strategies, including stimuli-responsive materials with programmable geometry and compliance, contactless manipulation modalities that reduce mechanical damage, reconfigurable swarm formations, and remotely applied magnetic, acoustic, and optical fields \cite{chen2025physical}. Across these approaches, a common objective is to leverage dominant surface-mediated interactions while remaining within the narrow force tolerances of delicate biological targets.

\subsection{Definition and Capability Space of Micro-Dexterity}


We define ``micro-dexterity" as the capability of micromanipulation systems to construct, modulate, and adaptively regulate mechanical interactions with target objects in order to reliably alter their state under microscale conditions. In contrast to macroscale dexterity, which relies on fully actuated multi-DOF end-effectors and rich proprioceptive feedback, micro-dexterity is governed by low-Reynolds-number dynamics, adhesion-dominated interactions, limited direct sensing, and often globally coupled actuation. These constraints become especially stringent when handling soft biological matter, for which force limits, viability preservation, and target heterogeneity further restrict feasible manipulation strategies. The operational environment, spanning from open substrates through microfluidic confinements to in vivo vasculature, actively mediates these constraints by determining viable strategies and available feedback. Accordingly, micro-dexterity should be evaluated not only by the breadth of executable primitives, but also by the fidelity of force regulation, adaptability to target variability, composability across multi-step tasks, and robustness under stochastic perturbations.

We organize the ``micro-dexterity"  capability domain into a multi-dimensional space defined by three axes, as illustrated in Fig. \ref{fig:overview}: manipulation primitives, target objects, and operational modes.

\textbf{\textit{Manipulation primitives:}} The fundamental atomic actions that constitute complex tasks, including grasping and holding, reorienting, precise releasing or placing, and assembling. In specialized contexts such as microsurgery and intracellular operation, these extend to penetrating, anchoring, probing, cutting, or other task-specific maneuvers.


\textbf{\textit{Target objects:} }The items being manipulated range from synthetic micro-objects such as microbeads and hydrogel constructs to biological structures including single cells, spheroids, organoids, and tissue constructs. Their size, stiffness, and fragility define the permissible force envelope and contact strategy. 
A high level of micro-dexterity is reflected in the capacity to accommodate this range of objects without relying on fundamentally different robotic platforms.


\textbf{\textit{Operational modes:} }The strategy by which manipulation is achieved, including contact versus contactless approaches, constrained (grasping or enclosing) versus unconstrained (pushing or steering) interactions, and single agent versus multi-agent or swarm coordination. Mode selection depends on task requirements and the operational environment, from open substrates through microfluidic confinements to in vivo vasculature, balancing gentle non-contact handling against the more precise spatial control afforded by direct contact.


\section{Embodiments for Micro-Dexterous Manipulation}

\begin{figure*}
    \centering
    \captionsetup{font=footnotesize,labelsep=period}
    \includegraphics[width=\linewidth]{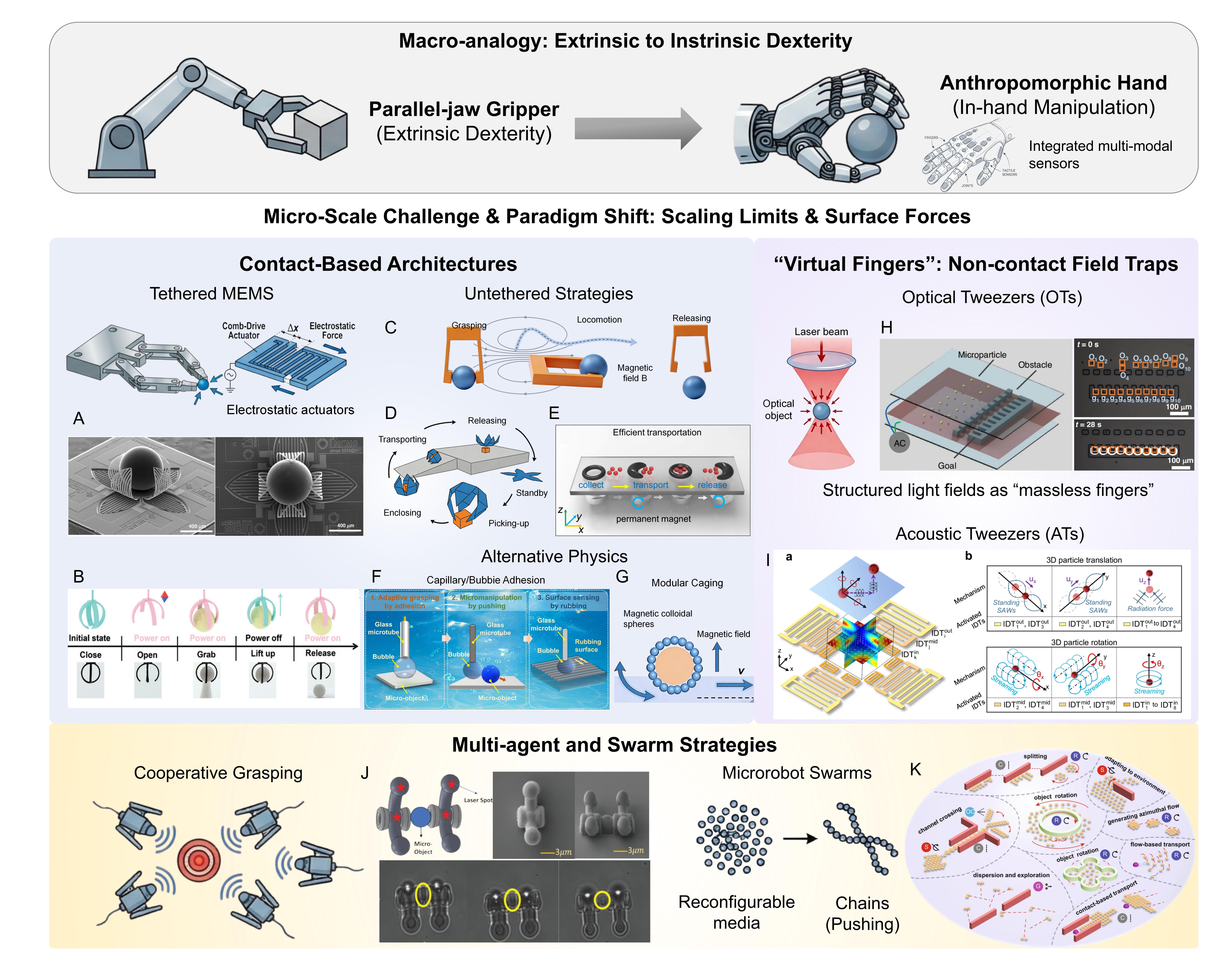}
    \caption{Overview of embodiments for micro-dexterous manipulation: harnessing diverse physical interactions and scaling laws to engineer embodied intelligence at the microscale. Tethered MEMS actuators: (A) an electrothermally actuated bimorph microgripper and (B) an electrothermal-magnetic actuated shape memory microgripper. Untethered strategies mirroring the macro-evolution from simple tweezers to complex multi-finger enclosing architectures: (C) a two-finger magnetic microgripper for grasp–transport–release and (D) a multi-arm microgripper with programmable 3D magnetization, (E) reconfigurable multifunctional ferrofluid droplet robots. Alternative physics approaches exploiting (F) capillary forces in bubble-based manipulation and (G) rolling entrapment via magnetic lassos. ``Virtual Fingers" enabling non-contact dexterity via field traps: (H) collision-free multi-object navigation via OET and (I) multi-DoF manipulation via acoustic traps. (J) Distributed 6D manipulation via multi-agent and (K) reconfigurable magnetic swarms for collective tasks. Panel (A) is adapted from Ref.~\cite{yang2024microgripper}, CC BY 4.0; Panel (B) is reprinted with permission from Ref.~\cite{Wu2025High‐Load}, Copyright 2025, Wiley-VCH GmbH; Panel (E) is reprinted from Ref.~\cite{fan2020reconfigurable}, CC BY-NC-ND 4.0; Panel (F) is reprinted from Ref.~\cite{xu2025bubbles}, CC BY 4.0; Panels (H), (I), (J), and (K) are adapted from Refs.~\cite{zheng2025automated}, \cite{qin2024movable}, \cite{zhang2020distributed}, and \cite{gardi2022microrobot}, respectively, all under CC BY 4.0.}
    \label{fig:embodiment}
\end{figure*}



In the macroscale domain, robotic manipulation has evolved from simple parallel-jaw grippers to multi-fingered anthropomorphic hands that achieve intrinsic dexterity by coordinating multiple articulated digits for in-hand manipulation \cite{yu2022dexterous, kadalagere2023review}. These systems routinely integrate tactile arrays, torque sensors, and vision to infer contact states and adapt grip \cite{li2024comprehensive, cong2024tacfr}. 
However, the direct translation of this architecture to the microscale is hindered by constraints in fabrication and sensor integration, together with the physical predominance of surface forces over inertia.
 Consequently, micro-dexterity demands a paradigm shift: replacing rigid joints with compliant mechanisms or field-mediated manipulation, and compensating for the absence of direct force sensing with indirect visual or deformation feedback. The following sections explore these embodiments, categorized by their interaction modality: physical contact, non-contact fields, and collective manipulation.

Manipulation at the microscale presupposes that the actuating agent, whether a mobile microrobot or an externally projected field, has been positioned or directed to engage the target. Magnetic fields exploit torques and gradient forces for propulsion and steering, offering deep tissue penetration and high biocompatibility \cite{hussein2023actuation}. Acoustic fields generate radiation forces and streaming flows that enable contactless thrust. Optical fields provide sub-nanometer spatial resolution for localized trapping and positioning. Chemical and biohybrid strategies achieve autonomous propulsion without external field infrastructure \cite{hou2023review}. Beyond locomotion, these fields also mediate the physical interactions that underpin object manipulation. As illustrated in Fig. \ref{fig:embodiment}, the following sections examine how embodiment designs and field-mediated strategies convert such actuation into interaction with microscale targets.


\subsection{Contact Manipulation via Engineered End-effectors}


Contact-based methods physically constrain the target via jaws or by exploiting cell–surface interactions and remain the mainstream solution. Analogous to miniaturized robotic hands, fingered tools have evolved from simple cantilever tweezers to sophisticated multi-fingered end-effectors.

Among these, microelectromechanical (MEMS)-based microgrippers represent a highly mature paradigm, leveraging semiconductor fabrication processes to achieve output forces and integrated feedback. Electrostatic actuators, typically employing comb-drive architectures, offer rapid response times and low power consumption, though often with limited output force \cite{algoos2024electrostatically}.
Piezoelectric systems, often coupled with flexure hinges for displacement amplification, offer superior bandwidth and sub-micron motion resolution \cite{guo2023design}. Electrothermal mechanisms utilize Joule heating and differential thermal expansion to generate large-stroke displacements, balancing structural simplicity with high work output \cite{yang2024microgripper, hussain2025electrothermally}.
Shape-memory-alloy (SMA) microgrippers leverage temperature-induced phase transformations, achieving high power-to-weight ratios and large latching forces  \cite{Wu2025High‐Load}. A distinct advantage of these tethered architectures is the capacity for integrated proprioception; capacitive or piezoresistive sensors embedded within the structural frame facilitate closed-loop force control, allowing for the gentle handling of fragile biological samples without crushing.

Untethered end-effectors eliminate physical tethers to external hardware, enabling operation in confined or enclosed environments. Soft polymer two-finger tweezers with embedded magnetic particles or anisotropic magnetization enable wireless opening/closing under uniform fields \cite{diller2014three,zhang2024enhanced}, and origami-inspired  multi-finger architectures expand the range of stable and encircling grasps \cite{xu2019millimeter}. Lacking onboard motors, these end-effectors rely on smart material mechanics and external fields for actuation. Thermal actuators typically exploit the mismatch in thermal expansion coefficients between layered materials or the volume phase transitions of hydrogels to drive large-stroke deformation \cite{danielson2017fabrication}. Magnetically actuated designs utilize field-induced torques on embedded particles to drive shape transformations, ranging from the bending of discrete magnetized polymer segments for on-demand grasping to the morphological reconfiguration of ferrofluid droplets for payload \cite{shao20213d, zhao2023design,fan2020reconfigurable}. Light-responsive systems leverage spatially selective photon absorption to trigger localized photothermal or photochemical actuation, offering high addressability but typically limited output force \cite{gao2024light}. Hybrid microgrippers overcome single-mode limitations by integrating multiple modalities, exemplified by systems combining magnetic guidance with thermal shape-locking to significantly improve load capacity \cite{hu2022multifunctional,ahmad2023hybrid}. These hybrid schemes broaden the operational envelope of contact-based microgrippers in terms of force, speed, and task adaptability.

Complementing these grippers are alternative strategies that exploit unique microscale physics for gentler handling. Capillary gripping utilizes liquid bridges whose surface tension secures micro-objects without mechanical clamping \cite{kim2025nanoporous}. Bubble-based approaches employ microbubbles as compliant end-effectors that leverage interfacial adhesion for adaptive grasping and rotation in aqueous environments \cite{xu2025bubbles}. Geometric enclosure strategies use multiple field-driven agents to collectively constrain targets without direct compression, as demonstrated by spinning microrobot collectives harnessing fluidic torque for contactless object manipulation \cite{ceron2026fluidic}.


\subsection{Contactless Manipulation via Shaped Fields}

In contrast to contact-based end-effectors, non-contact methods generate ``virtual fingers” through shaped energy fields, manipulating objects without tangible material interfaces. Optical tweezers (OTs) and their variants utilize structured light fields to act as ``massless fingers", enabling the stable contour-tracking manipulation of large irregular particles \cite{zhang2022fabrication, omine2024manipulation}, automated collision-free navigation of multiple micro-objects \cite{zheng2025automated}, and versatile hybrid handling via coupled electro-optical traps \cite{gonzalez2025hybrid}. Magnetic tweezers (MTs) exploit localized field gradients from electromagnet arrays to trap and manipulate functionalized magnetic beads with piconewton force resolution, enabling intracellular positioning and mechanical characterization of subcellular structures \cite{wang2019intracellular, miyamoto2026high}. For larger or opaque targets, acoustic tweezers (ATs) employ ultrasonic standing waves generated by phased arrays to create movable pressure nodes. These acoustic fields exert stronger forces and function as reconfigurable ``virtual hands", allowing for the parallel manipulation of multiple objects \cite{qin2024movable,li2025petri,shen2024joint}.

\subsection{Hybrid and Cooperative Strategies}
While collective strategies can employ both contact-based and non-contact interaction modalities, we treat them as a distinct embodiment because dexterity arises primarily from coordination among agents rather than from the design of any individual component.
Cooperative trapping uses multiple independent agents to constrain a target from opposing sides, thereby establishing stable force closure \cite{zhang2020distributed,barbot2019floating}.
Microrobot swarms act as reconfigurable media that exploit morphological plasticity to manipulate targets, dynamically transforming into functional architectures such as chains or ribbons for directional pushing, hydrodynamic vortices for non-contact trapping \cite{xie2019reconfigurable, gardi2022microrobot}, or self-assembling into tubular structures to enclose and compress cargo \cite{wang2023colloidal}. This structural versatility enables the robust transport of heterogeneous payloads.


\section{Sensing, Perception, and State Estimation for Micro-Dexterous Manipulation}

\begin{figure*}
    \centering
    \captionsetup{font=footnotesize,labelsep=period}
    \includegraphics[width=1\linewidth]{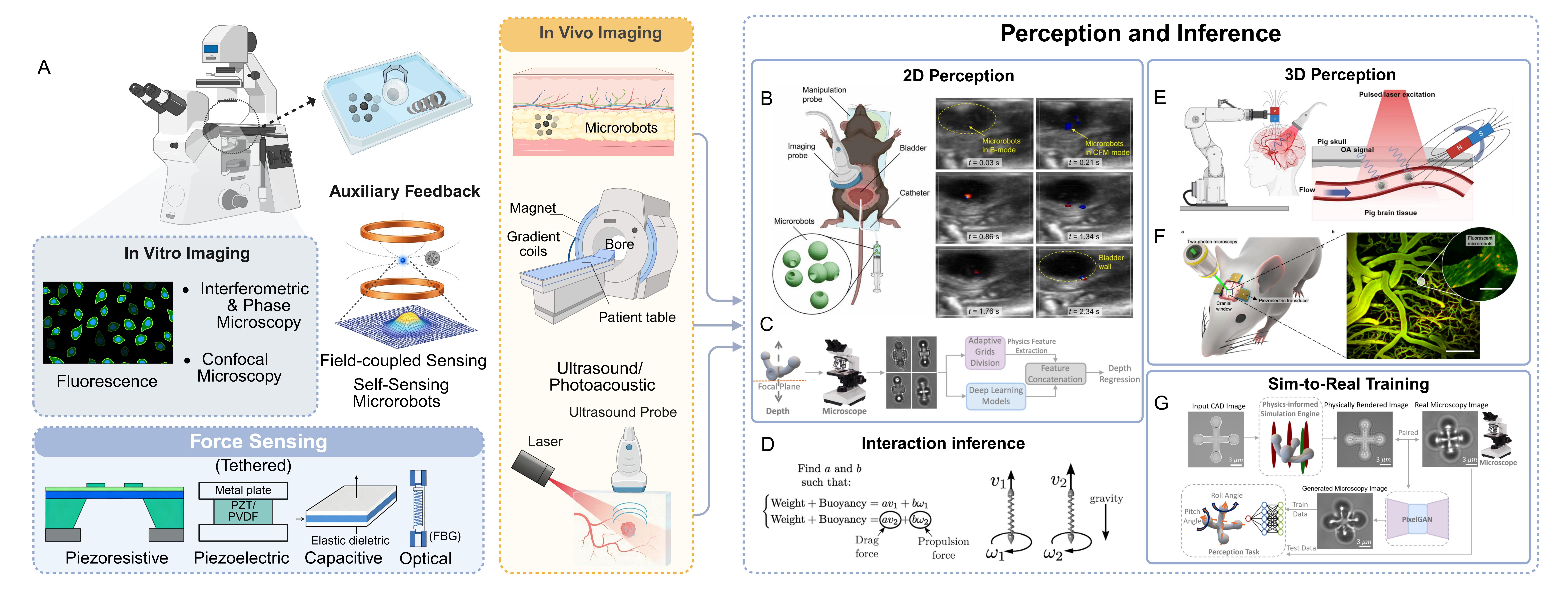}
    \caption{Sensing, perception, and state estimation for micro-dexterous manipulation. (A) Overview of multi-modal sensing modalities, including tethered force sensors, in vitro and in vivo imaging, and field-coupled sensing. (B-C) Computational pipelines for planar localization, tracking, and depth inference. (D) Force inference from observed kinematic responses. (E-F) Volumetric state estimation for deep-tissue navigation via optoacoustic tracking and vascular mapping. (G) Sim-to-real frameworks for training perception models under limited experimental data. Panel (B) is reprinted from Ref.~\cite{dillinger2025real}, CC BY 4.0; Panel (C) is reprinted from Ref.~\cite{wei2025physics}, CC BY 4.0; Panel (D) is reprinted with permission from Ref.~\cite{barbot2017helical}, Copyright 2017, Elsevier; Panel (E) is reprinted from Ref.~\cite{wang2025optoacoustic}, CC BY 4.0; Panel (F) is reprinted from Ref.~\cite{del2023ultrasound}, CC BY-NC-ND 4.0; Panel (G) is adapted from Ref.~\cite{tan2025physics}.}
    \label{fig:perception}
\end{figure*}

Micro-dexterous interaction relies on a sensing-to-perception pipeline: sensing refers to the physical transduction of quantities such as force or displacement into measurable signals, while perception denotes the computational extraction of actionable state estimates from raw sensory data. At the microscale, this pipeline is constrained by the scarcity of onboard sensing. Unlike macroscale hands that leverage dense tactile arrays and force/torque sensors \cite{li2024comprehensive}, micro-manipulators often face a trade-off between sensor integration and device scaling. Tethered systems can embed MEMS-based tactile sensors for quantitative force feedback, whereas untethered systems increasingly rely on computational vision and morphology-driven proxies \cite{adam2024overview}, necessitating an architecture that spans physical transduction, computational imaging, and physics-informed state estimation. Fig. \ref{fig:perception} provides an overview of these sensing and perception modalities.



\subsection{Contact Force Sensing: Physical Transduction Mechanisms}

For tethered manipulators, physical sensors rely on distinct mechanisms to convert interaction forces into readable electrical or optical signals. Piezoresistive sensors commonly utilize metal-foil or semiconductor strain gauges in wheatstone bridges and are favored for their simple readout circuits. However, detecting sub-nanonewton forces typically requires compliant mechanical designs, stress-concentration features, or mechanical amplification structures \cite{aubeeluck2023screen, chen2023nonlinearity}. In contrast, piezoelectric sensors, such as PZT or PVDF, operate without external excitation and offer high dynamic response suitable for detecting cell puncture events; however, inherent charge leakage limits their capacity for long-duration static force measurement \cite{sawane2023mems, he2024piezoelectric}. Alternatively, capacitive sensors achieve high sensitivity and thermal stability by measuring gap or area variations between electrode plates \cite{mishra2021recent}, where recent bio-inspired designs like swallow-tail structures have successfully decoupled mechanical and electrical interference to improve linearity \cite{gao2023bionic}. 
Finally, optical sensors such as fiber Bragg gratings (FBGs) and Fabry–Pérot interferometers offer ultra-high resolution and immunity to electromagnetic interference, although they remain susceptible to mechanical cross-coupling \cite{zhu2023advances}.

\subsection{Visual Perception: Multi-Scale Imaging Modalities}

Micro-dexterous manipulation often relies on visual perception for navigation and indirect force estimation; however, this multiscale imaging foundation requires a critical trade-off among resolution, penetration depth, and real-time capability \cite{gao2024intelligent}. Current sensing systems for micromanipulation can be broadly categorized into two domains: in vitro and in vivo.

In vitro settings are typically structured and accessible, where optical microscopy remains the primary feedback modality. Fluorescence wide-field imaging enables high-throughput observation when labels are available \cite{he2023high}. 
For label-free manipulation of optically transparent objects, phase-based contrast methods or interferometric phase microscopy are often required to recover morphological features and dynamic behavior \cite{nguyen2022quantitative}. 
Across optical microscopes, limited depth of field, out-of-focus background in thick samples, and diffraction-limited blur remain practical challenges, 
which complicate reliable pose estimation and contact inference at the microscale \cite{prakash2023super}. 
Optical sectioning techniques such as confocal microscopy suppress out-of-focus light. Confocal imaging has also been extended to multiview and super-resolution regimes, although these advances typically come at the cost of greater acquisition and reconstruction complexity \cite{wu2021multiview}.



For microrobots operating inside living organisms, optical approaches are frequently constrained by limited penetration depth, motivating the use of non-optical modalities for deep-tissue tracking. Magnetic resonance imaging (MRI) offers excellent penetration but suffers from slower refresh rates and lower spatial resolution than optical microscopy in microrobotic tracking \cite{li2024human}. Ultrasound imaging provides high temporal resolution for tracking fast-moving microrobots in vasculature but is limited by artifacts at bone or air interfaces \cite{del2023ultrasound}. Photoacoustic imaging has emerged as a hybrid imaging modality that integrates optical contrast with ultrasonic penetration, thereby enabling the visualization of microrobots in deep tissue with a balanced spatiotemporal resolution \cite{wang2025optoacoustic}.


\subsection{State Estimation for Untethered Manipulation}

Current approaches to contact state estimation for untethered microrobots typically depend on indirect inference, either by analyzing visually observed deformations of soft targets or through model-based estimation. In these systems, raw imaging data must be transformed into state estimates through computational processing pipelines. More recently, the integration of deep learning with physics-based imaging models has begun to alleviate the intrinsic limitations of micro-optical systems.

Accurate planar localization is fundamental to autonomous micro-control, relying primarily on intensity-based segmentation, morphological extraction, and feature tracking. Thresholding techniques, ranging from global binarization to adaptive schemes \cite{bendkowski2021autonomous}, are computationally efficient but can be sensitive to illumination inhomogeneity and imaging artifacts. Learning-based detection and tracking pipelines have therefore been increasingly adopted for robust planar localization in challenging microscopy scenes \cite{sawhney2024motion, an2024microscopic}. However, the reliability of these 2D methods is constrained by their dependence on high-quality focal planes, rendering them vulnerable to errors when targets undergo significant deformation or move out-of-plane.

Traditional axial depth ($Z$-depth) estimation in microscopy commonly relies on depth-from-focus (DFF) or depth-from-defocus (DFD) cues \cite{fan2025optical}. However, these cues become unreliable for transparent, texture-less microrobots. To address this limitation, data-driven methods have learned depth-sensitive representations from focus measures using regression models for microscopic depth and pose estimation \cite{zhang2020data}, while more recent physics-informed networks explicitly incorporate optical focusing cues to better capture the nonlinear relationship between axial position and image appearance \cite{wei2025physics}. Their performance, however, remains constrained by the scarcity of annotated microscopy data, since accurate depth labeling typically requires specialized experimental hardware \cite{wei2025dataset}. Sim-to-real strategies therefore play a central role by combining physics-based simulation with generative models to synthesize training data that better preserve depth-dependent optical cues, including focal-state-dependent spectral variations \cite{zhang2022micro}.


Beyond depth, full 3D pose estimation is essential for manipulation tasks that require precise alignment or reorientation. Data-driven approaches have jointly estimated microscopic depth and pose by combining focus-measure features with deep residual networks incorporating domain-specific prior knowledge \cite{zhang2020data}. Related learning-based autofocus frameworks have also been extended to 3D pose identification of micro- and nanowires in dynamic suspensions \cite{song2024learning}.  Nevertheless, reliable real-time 6-DoF estimation remains challenging for transparent or deformable micro-objects, whose image appearance varies substantially with orientation, focal offset, and optical interference, often without stable texture cues. These limitations highlight the need for perception frameworks that explicitly account for microscopy image-formation physics rather than relying solely on appearance-driven inference \cite{tan2025physics}.

For dexterous manipulation, estimating the interaction forces and torques between the actuating agent and the target is critical for closed-loop regulation and preventing damage to fragile biological samples. OTs quantify piconewton forces and piconewton-nanometre torques by correlating trapped particle displacement with trap stiffness and analyzing angular momentum transfer \cite{gerena20233d}, with optically driven chiral microstructures recently serving as tunable torque clamps on individual molecular motors \cite{donini2026optically}. In the acoustic domain, joint subarray tweezers infer both translational forces and viscous torques from controlled cell motion \cite{shen2024joint}, and inverse-designed metasurfaces have achieved simultaneous programmable force and torque control \cite{liu2025end}.



\subsection{Auxiliary Sensing Modalities for Manipulation Feedback} 

Beyond imaging and contact-based force transduction, additional sensing modalities provide task-specific feedback that directly informs manipulation decisions. Field-coupled methods measure the actuation fields themselves: magnetic particle imaging generates millimetre-scale 3D maps of superparamagnetic tracers for agent localization \cite{ahlborg2025invited}, while photoacoustic-robotic integration enables depth-resolved tracking of circulating agents through oscillatory acoustic signatures \cite{peng2025photoacoustic}. Self-sensing microrobots equipped with resonant circuits can wirelessly report local environmental properties by modulating the enhancement of external radio frequency fields \cite{li2023self}. At the task level, electrical and biochemical signals provide manipulation-specific feedback. Electrical impedance changes at the cell-pipette interface detect contact establishment and pre-detachment oscillations, enabling autonomous regulation of suction forces during micro-extraction \cite{dinca2025preliminary}. Fluorescence indicators fused with force sensing detect sublethal cellular injury during robotic single-cell surgery, allowing adjustment of manipulation strategies to preserve viability \cite{shakoor2022advanced}.

\section{Control and Learning for Micro-Dexterous Manipulation}

\begin{figure*}
    \centering
    \captionsetup{font=footnotesize,labelsep=period}
    \includegraphics[width=\textwidth]{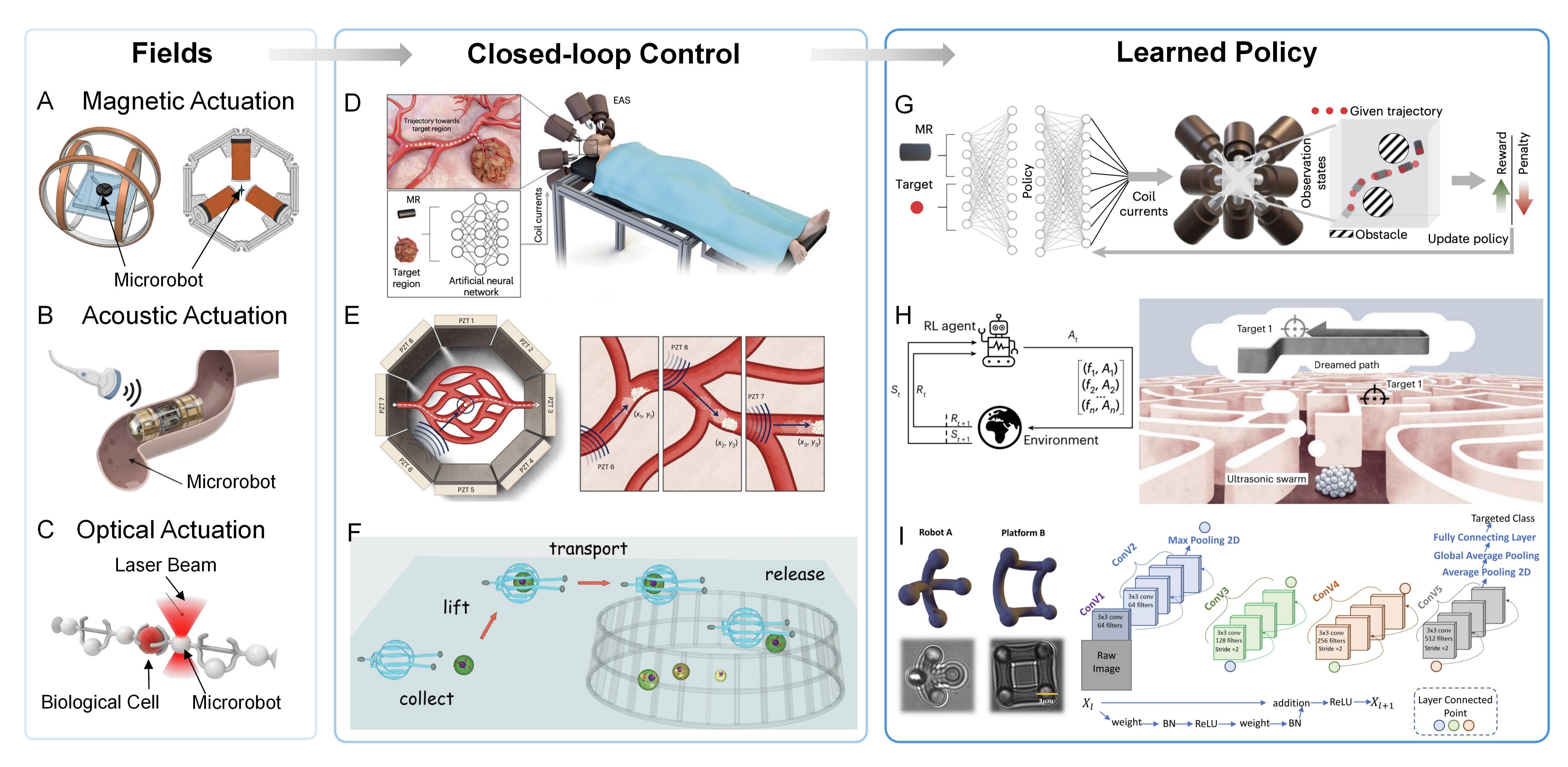}
    \caption{Control and learning for micro-dexterous manipulation. Columns show the progression from external field generation to closed-loop control and learned policies, while rows correspond to magnetic (A, D, G), acoustic/ultrasonic (B, E, H), and optical (C, F, I) actuation. (A-C) Representative actuation mechanisms for each field modality. (D-F) Perception-guided feedback control for manipulation in complex environments. (G-I) Learning-based policies that map observations to control actions for adaptive navigation and task execution. Panels (D) and (G) are reprinted from Ref. \cite{abbasi2024autonomous}, CC BY 4.0; Panels (E) and (H) are reprinted from Ref. \cite{medany2025model}, CC BY 4.0; Panel (F) is reprinted from Ref. \cite{ivanyi2024optically}, CC BY-NC 4.0; Panel (I) is reprinted from Ref. \cite{zhang2020data}, CC BY-NC 4.0.}
    \label{fig:control_learning}
    \vspace{-4pt}
\end{figure*}


In macroscopic robotics, dexterous manipulation typically relies on high-fidelity actuators, proprioceptive sensing, and rigid-body dynamic models, with control formulated through the precise regulation of joint torques to achieve force and position objectives. In contrast, micromanipulation operates under low-Reynolds-number dynamics, pronounced surface effects, and limited onboard sensing and computation, making externally applied fields the dominant means of actuation \cite{hussein2023actuation}. Therefore, microscale manipulation is often globally actuated and strongly coupled, such that low-dimensional control inputs govern high-dimensional system states, particularly in swarm-based or multi-DoF manipulation scenarios.


This section reviews control and learning strategies for micro-dexterous manipulation under physical constraints, covering deterministic field actuation, perception-guided feedback control, and emerging data-driven methods. We highlight a broader transition from analytical model-based control toward learning-enabled control, where reinforcement learning and imitation learning are increasingly used to address nonlinear dynamics, partial observability, and sim-to-real transfer, thereby advancing microrobotic autonomy. Fig. \ref{fig:control_learning} illustrates representative examples across these three control paradigms.

\subsection{From Global Actuation to Local Manipulation}

The central challenge in microscale control is the disparity between global, low-dimensional actuation inputs and the local, high-dimensional objectives required for dexterous manipulation. Magnetic, acoustic, and optical fields each offer distinct control trade-offs: magnetic fields provide deep penetration and biocompatible torque/gradient actuation but decay rapidly with distance \cite{hou2023review}. Acoustic fields generate stronger volumetric forces and enable multi-degree-of-freedom trapping including translation, rotation, and deformation of single cells \cite{shen2024joint}. Optical fields offer sub-nanometer positioning resolution and precise localized trapping, enabling high-fidelity single-particle control \cite{omine2024manipulation}. However, most field-mediated modalities apply forces globally, coupling the motion of all agents within the workspace. Even focused approaches such as optical trapping face scalability limits when independent control of multiple objects is required.

Selective manipulation therefore depends on breaking the symmetry inherent in global actuation, whereby a common input would otherwise drive all agents in a similar or coupled manner. One mainstream approach is to exploit heterogeneity in agent properties, such as differential frequency responses or anisotropic magnetization profiles, so that identical global inputs produce distinct responses across agents \cite{wang2023selective}.
Alternatively, spatially structured fields created by localized coil arrays or hybrid magneto-optical actuation enable independent 3D control of multiple collectives within the same workspace \cite{Sun2024Multiple}. Beyond engineered heterogeneity, environmental interactions, including wall effects, fluid dynamics, and collision-induced self-organization, can transduce low-dimensional global inputs into high-dimensional collective behaviors \cite{hao2022controlling}. These symmetry-breaking strategies are foundational to achieving independent agent coordination and cooperative manipulation at the microscale.

\subsection{Closed-Loop Control with Exteroceptive Feedback}

To mitigate stochastic microscale perturbations such as Brownian motion and flow disturbances, manipulation systems must transition from open-loop actuation to closed-loop feedback control. Visual servoing is the primary control paradigm for micro-manipulation, where microscopy-based state estimates are fed into controllers that regulate motion in real time. Classical approaches employ PID regulation and Kalman filtering for trajectory tracking under moderate disturbances \cite{jiang2022control, huang2019visual}. Model predictive control (MPC) extends these capabilities by incorporating system dynamics and actuation constraints into a receding-horizon optimization, enabling anticipatory compensation for nonlinear field-object interactions and environmental perturbations \cite{jiang2022control}. Active disturbance rejection control provides a model-free alternative by estimating and compensating lumped disturbances, which is advantageous given the uncertain nature of microscale dynamics \cite{jia2024efficient}.



For deep-tissue applications where optical feedback is unavailable, control systems rely on penetrating medical imaging modalities. Ultrasound provides real-time tracking with high temporal resolution for dynamic monitoring of microrobots in vasculature \cite{Pane2022Ultrasound, Han2024Imaging-guided}. Photoacoustic imaging combines optical specificity with ultrasonic depth for tracking in opaque tissues \cite{Wrede2022Real-time}. MRI and MPI enable navigation in deep-seated organs but with slower refresh rates that constrain control bandwidth \cite{Go2022Multifunctional, Xing2025TriMag}. The low spatiotemporal resolution of medical imaging relative to optical microscopy imposes limits on achievable closed-loop bandwidth, requiring controllers that are robust to delayed and sparse feedback.


Swarm control extends beyond single-agent regulation to the modulation of collective shape, density, and morphology. The feedback variables transition from individual coordinates to statistical descriptors of the swarm configuration, and distributed control algorithms facilitate autonomous navigation and multi-target transport in obstacle-dense environments \cite{An2024Model-Free}. Under hybrid magnetic-optical actuation, multiple collectives have achieved independent 3D locomotion and cooperative multitasking \cite{Sun2024Multiple}.

\begin{figure*}[!t]
  \centering
  \captionsetup{font=footnotesize,labelsep=period}
  \includegraphics[width=1\textwidth]{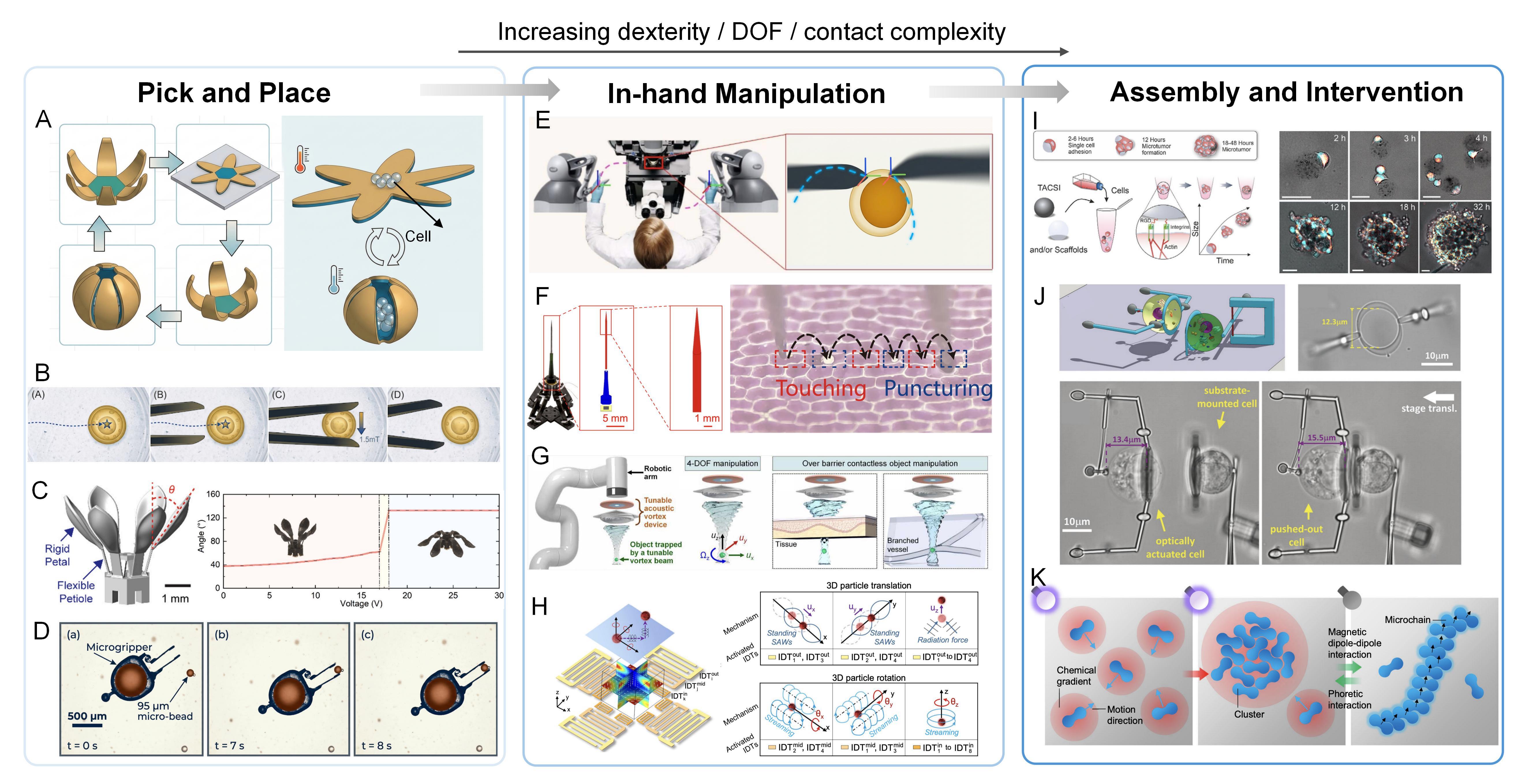}
  \caption{Task-level micro-dexterous capabilities arranged by increasing dexterity complexity. (A-D) Pick-and-place: reversible capture, transport, and release using soft encapsulation and programmable microgrippers. (E-H) In-hand manipulation: reorientation enabled by micromanipulators and field-defined ``virtual fingers". (I-K) Assembly and intervention: micro-assembly, probing and interventional actions, and collective behaviors for cooperative transport and construction. Panel (C) is reprinted from Ref. \cite{shao20213d}, CC BY-NC-ND 4.0; Panel (F) is reprinted from Ref. \cite{feng2025miniature}, CC BY-NC-ND 4.0; Panel (G) is reprinted from Ref. \cite{li2024robot}, CC BY-NC 4.0; Panel (H) is reprinted from Ref. \cite{shen2024joint}, CC BY-NC-ND 4.0; Panel (I) is reprinted from Ref. \cite{harder2025soft}, CC BY 4.0; Panel (J) is reprinted from Ref. \cite{ivanyi2024optically}, CC BY-NC 4.0; Panel (K) is reprinted from Ref. \cite{urso2023reconfigurable}, CC BY 4.0.}
  \label{fig:micro_dexterity}
  \end{figure*}

\subsection{Data-driven and Learning-based Strategies}

The complexity of microscale physics, characterized by fluid-structure interactions, uncertain boundaries, and non-Newtonian behavior, precludes precise analytical modeling, favoring data-driven approaches. Mirroring macrorobotic trends in solving complex contact dynamics, microscale RL is evolving beyond navigation toward manipulation.

Model-free RL approaches such as soft actor-critic (SAC) and trust region policy optimization (TRPO) enable microrobots to learn efficient swimming gaits in non-Newtonian fluids and achieve robust path tracking without prior fluid mechanics knowledge \cite{abbasi2024autonomous,Wang2025Deep}. End-to-end policy learning bypasses intermediate modeling by directly mapping visual inputs to actions \cite{Hao2025Learn-Gen-Plan:}. 
Multi-agent RL (MARL) facilitates swarm coordination for collaborative tasks such as distributed transport, with counterfactual reward designs promoting individual contributions to collective performance \cite{heuthe2024counterfactual, zhu2025lamarl}.

Beyond locomotion, learning-based methods have been increasingly extended to manipulation tasks that involve physical interaction and multi-step action sequencing. For example, a scheduling-adaptive imitation learning framework has been used to coordinate dual micropipettes for membrane stripping and embryo rotation, which achieves human-level success rates in long-horizon manipulation of deformable cells \cite{11230522}.
For OTs-driven systems, the Interactive OT Gym provides a high-fidelity digital twin platform for RL-based training of manipulation policies \cite{Tan2025Interactive}. At the macroscale, human-in-the-loop RL has demonstrated impressive performance on dexterous tasks including precision assembly and dual-arm coordination, learning near-perfect policies within hours of real-world training \cite{luo2025precise}. These advances suggest a viable pathway for transferring RL strategies to the microscale.


Despite this progress, a significant disparity persists between analytic field control and learning-based methods at the microscale. Field-based approaches have realized sophisticated manipulation  including bubble-based grasping \cite{xu2025bubbles}, distributed optical force control for coordinated multi-spot manipulation \cite{zhang2020distributed}, and acoustic subarrays for controllable cell rotation and deformation \cite{shen2024joint}. In contrast, microscale RL remains largely limited to locomotion and transport, lacking the capacity for complex reorientation, force-regulated grasping, or multi-step assembly \cite{barbot2019floating}. Bridging this gap demands high-fidelity simulation environments that capture microscale contact mechanics, reward formulations encoding manipulation quality beyond binary success, and scalable sim-to-real transfer accounting for domain-specific uncertainties. The convergence of digital twins, physics-informed learning, and adaptive field control will be essential to elevate microscale RL toward true dexterous manipulation.



\section{Task-Level Micro-Dexterous Capabilities in Biological Contexts}

Building on the micro-dexterity capability space established earlier, this section evaluates how current microscale manipulation systems instantiate these primitives in biological contexts, assessing actual performance and exposing gaps between laboratory demonstrations and clinically viable operation. We organize this evaluation around three representative tasks: pick-and-place, in-hand reorientation, and physical interaction for assembly and intervention, with representative demonstrations shown in Fig. \ref{fig:micro_dexterity}.

\subsection{Pick-and-Place: Capture, Transport, and Release}

Recent advances have extended biological cargo handling beyond passive delivery toward pick-and-place capability, characterized by reversible capture, controlled transport, and on-demand release of discrete targets. This primitive requires reliable target engagement, stable retention during motion, and deterministic release at a desired location. It is increasingly realized through diverse mechanisms including compliant microgrippers for single-cell capture, shape-morphing magnetic bodies that encapsulate living cargo, and collective assemblies that cooperatively enclose and transport micro-objects in confined environments \cite{zhang2024enhanced, xie2019reconfigurable, sun2022reconfigurable}.

At the capture stage, untethered soft grippers utilizing liquid crystal elastomers or hydrogels achieve on-demand grasping via external stimuli, enabling selective capture of single cells or tissue biopsies \cite{shao20213d, ahmad2023hybrid, kim2025nanoporous}. Magnetic soft microrobots integrating gripper function with locomotion in a single body enable simultaneous grasping and trajectory-tracked transport using three coil pairs \cite{wang2024magnetic}. For transport, magnetic slime robots dynamically alter their viscosity and morphology to form caging structures that envelop and carry objects through constricted channels \cite{sun2022reconfigurable}, while modular swarms cooperatively enclose cargo and transport it against physiological flows \cite{xie2019reconfigurable, jiang2024automated}. At the release stage, precise detachment remains a central challenge due to dominant surface adhesion at the microscale; strategies include stimuli-triggered shape reversal, controlled field reduction, and exploiting capillary force modulation for deterministic placement \cite{ yin2019untethered, kim2025nanoporous}.

Despite these proof-of-concept advances, a fundamental ``dexterity gap" still separates current microscale pick-and-place systems from macroscale dexterous manipulation. First, microscale systems are inherently underactuated: whereas macroscopic manipulators achieve arbitrary 6-DoF motion through fully actuated kinematic chains, microscale robots are typically driven by global fields that couple the motion of all agents and restrict articulation to low-dimensional control manifolds. Second, microscale manipulation remains constrained by limited sensing and control bandwidth: unlike macroscale platforms, which rely on high-rate multimodal feedback to regulate contact in real time, microscale systems generally lack onboard sensing and depend on external imaging with limited spatial and temporal resolution, thereby hindering rapid closed-loop response under physiological disturbances. Third, their functional repertoire remains narrow: whereas macroscopic robots can compose hierarchical, multistage manipulation strategies, most microscale systems remain task-specific and have limited capacity for sequential, reversible, and task-adaptive interactions in unstructured environments.


\subsection{In-Hand Manipulation and Reorientation}

Translating in-hand manipulation (IHM) primitives to the microscale requires replacing rigid mechanical fingers with specialized end-effectors or ``virtual fingers" generated by programmable fields. We use ``in-hand reorientation" specifically to denote the controlled orientation change of a payload relative to the actuator within a physical grasp or field-defined trap without release, as distinct from rigid-body rotation of the coupled actuator-payload assembly.

\textbf{Contact-based in-grasp reorientation.} In systems where the end-effector physically contacts the target, friction-driven rotation with robotic micropipettes has been used to reorient single cells such as oocytes for biopsy \cite{cui2020robotic}. A scheduling adaptive imitation learning framework recently coordinated dual micropipettes to strip membranes and rotate embryos with human-level success rates in long-horizon deformable manipulation \cite{11230522}. Origami-inspired micro-hands such as Micro-X4 mimic the kinematic redundancy of a macroscale hand, offering three-dimensional positioning and rotation with sub-micrometer precision for in-grasp reorientation during cell puncturing and injection \cite{feng2025miniature}. Multifunctional magnetic muscles combine phase-change polymers and magnetic particles to locally tune stiffness and shape, gently wrapping and reorienting fragile targets in a hand-like manner \cite{seong2024multifunctional}.

\textbf{Field-defined in-trap reorientation. }Non-contact approaches realize virtual fingers as programmable trap potentials, where the target is reoriented within the trap without physical contact from a solid end-effector. Chirality-tunable acoustic vortex tweezers use orbital angular momentum to roll trapped objects in 4 DOF, enabling tomographic inspection through overlying tissue \cite{li2024robot}. Joint subarray ATs extend this to controllable single-cell translation, rotation, and deformation, enabling in-trap reorientation while simultaneously probing mechanical responses \cite{shen2024joint}. In the optical domain, planar multi-spot OTs have demonstrated distributed force control, enabling coordinated multi-trap regulation of interaction forces and torques during manipulation \cite{zhang2020distributed}.

While these systems successfully emulate specific IHM primitives, they face distinct challenges compared to macroscopic dexterity. (i) Dominance of surface forces: unlike macro-IHM where gravity and friction allow predictable sliding and rolling, microscale reorientation must contend with strong adhesion and electrostatic interactions that make controlled relative motion between grasper and target difficult to sustain; (ii) Absence of interaction feedback: macro-IHM relies on high-frequency force and tactile sensing to detect incipient slip and estimate contact stability, whereas microscale systems depend on visual feedback with inherent latency and occlusion, preventing the reflex-based force modulation needed for targets with unknown compliance; (iii) Limited generalizability: macroscale multi-fingered hands manipulate diverse objects from rigid tools to soft fabrics, whereas current microscale reorientation systems are specialized for specific target geometries and cannot readily adapt across tasks.



\subsection{Physical Interaction: Assembly and Intervention}


This category encompasses two related but distinct task families unified by the requirement for sustained or sequential physical interaction with the target: structured micro-assembly, and active intervention or probing in unstructured biological settings. Assembly examples from general microsystems contexts are included where they reveal transferable primitives relevant to tissue engineering and biofabrication.

\textbf{Micro-assembly for functional systems. } In structured settings such as MEMS and tissue engineering, micro-assembly focuses on the precise integration of discrete components into functional architectures. Magnetic actuation, OTs, and ultrasound fields have enabled the manipulation of hydrogels, silica spheres, and carbon fibers to construct intricate micro-structures \cite{barbot2019floating, chizari2020automated, niendorf2021combining}. Notably, a floating microrobot pair enabled the precise wet transfer of planar films onto curved 3D substrates, effectively bridging planar fabrication with three-dimensional integration \cite{barbot2019floating}. Transitioning toward biological applications, externally actuated systems transport and position stem cells within micro-environments, while optical microrobots utilize light-based trapping for the high-precision sorting and patterning of single cells to engineer functional tissues \cite{hou2023review, harder2025soft}. These assembly tasks share common manipulation requirements including sub-micrometer relative positioning and stable multi-point contact, which are complicated by dominant surface adhesion and limited real-time sensory feedback at the microscale.

\textbf{Intervention and intracellular probing.} In unstructured in vivo environments, physical interaction translates to mobile agents navigating to a target and executing active mechanical work. The core manipulation demands shift from precise registration to target engagement, penetration, and anchoring. Helical nanorobots driven by rotating magnetic fields function as untethered mobile drills capable of mechanically penetrating live cell membranes and navigating to sub-cellular structures for payload delivery \cite{chen2022carbon}. At the tissue scale, theragrippers inspired by parasitic hooks serve as autonomous surgical clamps, mechanically anchoring onto mucosal tissues for sustained drug release \cite{zhao2023design, ghosh2020gastrointestinal}. Wireless micro-probes act as intracellular sensors, utilizing rotation to measure local cytoplasmic viscosity \cite{pal2020helical}. Unlike assembly in controlled settings, intervention must contend with unpredictable tissue mechanics and immune responses, demanding manipulation strategies robust to environmental uncertainty.

Assembly and intervention tasks are particularly demanding at the microscale. Independent tool coordination and error correction are limited by global field actuation, which couples the motion of all agents and makes it difficult to independently position components or correct misplacement during multi-step sequences. Contact transitions are governed by adhesion and surface effects rather than reversible mechanical fastening, making reliable clamping, controlled release, and repeatable component joining difficult during assembly. Sparse force and torque feedback limits force-aware interaction, making assembly alignment errors, penetration depth, and contact state transitions difficult to detect and regulate.


\begin{figure*}
    \centering
    \captionsetup{font=footnotesize,labelsep=period}
    \includegraphics[width=0.8\linewidth]{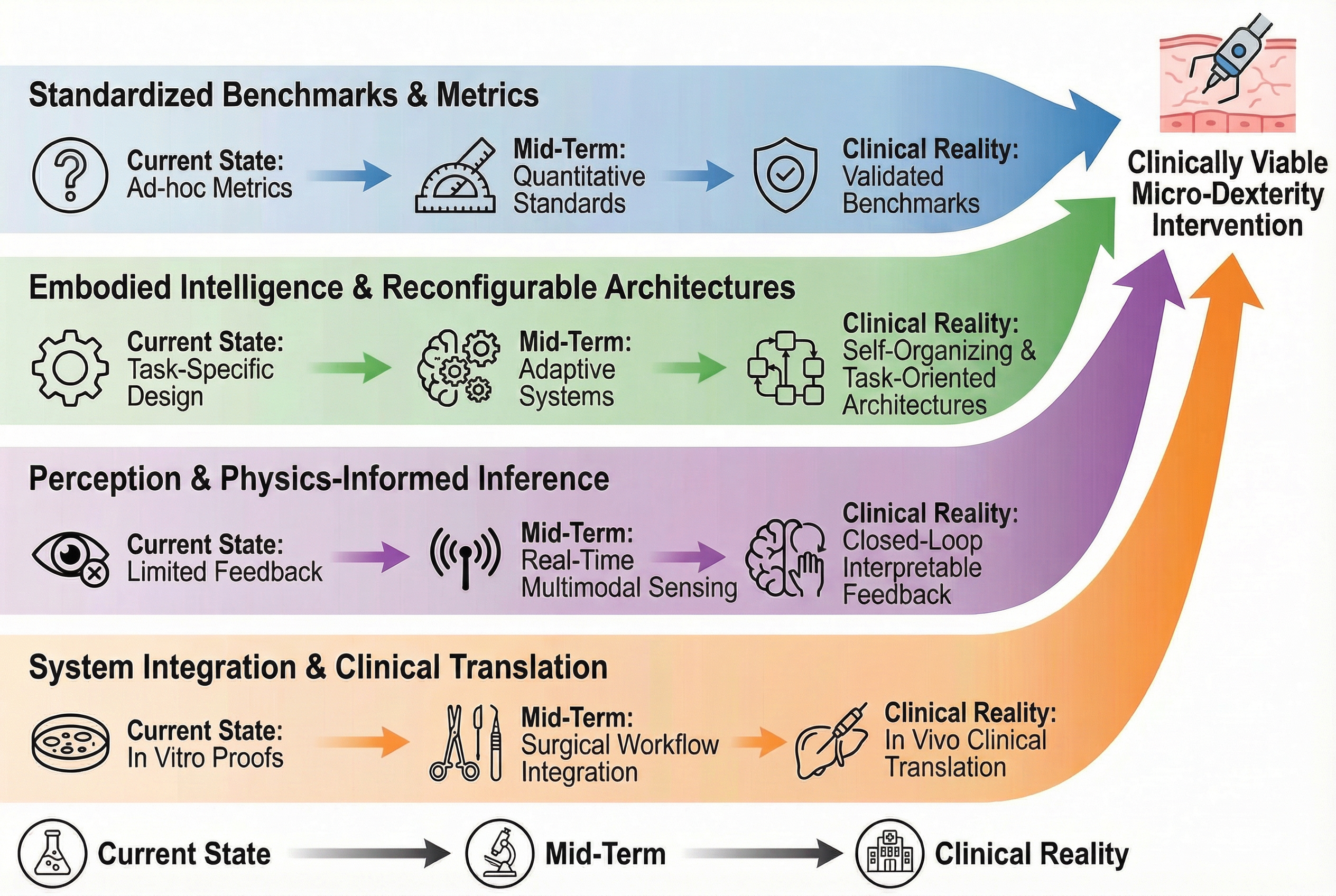}
    \caption{Roadmap towards clinically viable micro-dexterity. This chart illustrates the projected evolution of microrobotics to bridge the current ``dexterity gap" where four parallel tracks are outlined: (1) establishing standardized benchmarks and metrics; (2) shifting towards embodied intelligence and reconfigurable architectures; (3) closing the feedback loop via wireless sensing and physics-informed inference; and (4) translating system integration from in vitro proofs to compliant in vivo clinical interventions.}
     \label{fig:roadmap}
    \vspace{-5pt}
\end{figure*}

\section{OUTLOOK: TOWARDS CLINICALLY VIABLE MICRO-DEXTERITY}

Despite rapid progress, microscale manipulation remains largely at the stage of isolated demonstrations, including grasping, transport, and reorientation under controlled laboratory conditions. A persistent ``dexterity gap'' therefore separates current capabilities from the integrated, feedback-regulated, multi-step manipulation required for clinical intervention. As illustrated in Fig.~\ref{fig:roadmap}, closing this gap requires coordinated advances along four complementary axes: standardized benchmarks and metrics, embodied intelligence and reconfigurable architectures, perception and physics-informed inference, and system integration and clinical translation. Across all four axes, the developmental trajectory is similar: from the current state of proof-of-concept systems, through mid-term quantitative and workflow-level integration, toward clinically validated micro-dexterity intervention.

\textbf{Standardized benchmarks \& metrics.} Micro-dexterity demands high-fidelity force control, sub-micrometer positioning accuracy, and task versatility comparable to those of macroscopic manipulation, but under far stricter physical, sensing, and environmental constraints. Yet the current state of the field remains characterized by ad hoc evaluation, with most studies reporting task success only in platform-specific settings. While the macroscopic domain benefits from established benchmark suites such as YCB \cite{calli2015benchmarking}, analogous standards for micro-dexterity are still lacking. A key mid-term priority is therefore the definition of canonical task families, standardized micro-object sets, and quantitative metrics that capture placement and release accuracy, force regulation during contact-rich interaction, robustness to variability in target size and stiffness, multi-step task composability, and viability preservation for biological samples. The long-term goal is a validated benchmarking framework that enables reproducible cross-platform comparison, rigorous assessment of manipulation capability, and objective evaluation of clinical readiness.

\textbf{Embodied intelligence \& reconfigurable architectures.} At present, most systems remain task-specific by design, with embodiments optimized for a narrow manipulation primitive or a single operating environment. Moving beyond this current state requires adaptive systems that can reconfigure their morphology, interaction mode, or collective organization in response to task demands. In the mid-term, such adaptability should support transitions across locomotion, grasping, transport, release, and contact regulation without requiring a complete redesign of the platform. The long-term clinical objective is self-organizing and task-oriented architectures that can robustly coordinate multiple agents, interaction modalities, and manipulation primitives under physiological constraints. In this sense, clinically viable micro-dexterity will depend not only on better materials or actuators, but on embodied intelligence that couples mechanical design, control authority, and task adaptability.

\textbf{Perception, physics-informed inference, and control.} Limited feedback remains a central bottleneck for microscale dexterous manipulation. Unlike macroscale manipulators, which benefit from dense proprioception and tactile sensing, microscale systems typically rely on delayed visual feedback and only limited direct measurement of contact states. Closing the feedback loop in biological environments, particularly in deep tissue, therefore requires coordinated advances in sensing, state estimation, and control. On the perception side, future systems will need real-time multimodal sensing that combines microscopy, ultrasound, photoacoustic imaging, magnetic sensing, and other wireless modalities to improve observability across spatial scales and penetration depths. Because such measurements are often partial and indirect, physics-informed inference will be equally important for fusing imaging data with biomechanical and hydrodynamic models to estimate contact, force, and deformation in occluded and deformable environments. These perception and inference capabilities must in turn support more capable control frameworks. In particular, reinforcement learning policies trained in high-fidelity digital twins should evolve beyond navigation toward force-aware, contact-rich manipulation, enabling robust operation under the model uncertainty and unmodeled disturbances inherent to biological settings. Ultimately, the clinical goal is closed-loop, interpretable, and reliable feedback control that can support safe manipulation of fragile biological targets in heterogeneous tissue environments.

\textbf{System integration \& clinical translation.} Translating micro-dexterous manipulation into clinically viable tools requires overcoming system-level challenges that extend well beyond the performance of individual components. At present, the field is still dominated by in vitro proof-of-concept demonstrations, where impressive manipulation behaviors are achieved in simplified environments but remain largely disconnected from real clinical workflows. A critical mid-term step is therefore the integration of individual manipulation primitives into reliable task-level pipelines that operate alongside standard imaging modalities, microfluidic platforms, catheter-based delivery systems, and established biological protocols. This transition demands not only improved manipulation capability, but also compatibility with sterilization, navigation, targeting, user interfaces, and broader procedural constraints. Ultimately, in vivo clinical translation will require micro-dexterous systems that can function robustly in heterogeneous physiological environments while satisfying biocompatibility requirements. Clinical viability will depend not only on more capable devices, but also on reproducible manipulation workflows, robustness across operators and laboratories, compatibility with existing clinical hardware, and repeatable performance under regulatory scrutiny.

 Taken together, these four directions define a roadmap from the current state of fragmented demonstrations to clinically viable micro-dexterity intervention. In summary, the micro-dexterity framework proposed in this review argues that the path forward lies not in isolated optimization of materials, actuation, or sensing, but in their coordinated co-design around task-adaptive interaction capability. By anchoring future research to quantifiable dexterity levels, standardized benchmarks, and integrated system design, the field can systematically advance from current single-primitive demonstrations toward the robust, multi-step, feedback-regulated biological manipulation that clinical translation demands.   

\bibliographystyle{IEEEtran}  
\bibliography{ref_manipulation.bib}           

\end{document}